\documentclass[sigconf]{acmart}

\usepackage{amsmath}
\usepackage{tcolorbox}
\usepackage{graphicx}
\usepackage[table,xcdraw]{xcolor}
\usepackage{multirow}
\tcbset{colback=white, colframe=black, boxrule=0.5pt}
\AtBeginDocument{
  }

\setcopyright{acmlicensed}
\copyrightyear{2018}
\acmYear{2018}
\acmDOI{XXXXXXX.XXXXXXX}
\acmConference[Conference acronym 'XX]{Make sure to enter the correct
  conference title from your rights confirmation email}{June 03--05,
  2018}{Woodstock, NY}

\acmISBN{978-1-4503-XXXX-X/2018/06}

\begin{document}

\title{Cognitive-Aligned Spatio-Temporal Large Language Models For Next Point-of-Interest Prediction}

\author{Penglong Zhai}
\authornote{Contributed equally to this research.}
\affiliation{
  \institution{AMAP, Alibaba Group}
  \city{Beijing}
  \country{China}
}
\email{zhaipenglong.zpl@alibaba-inc.com}

\author{Jie Li}
\authornotemark[1]
\affiliation{
  \institution{AMAP, Alibaba Group}
  \city{Beijing}
  \country{China}
}
\email{lj313796@alibaba-inc.com}

\author{Fanyi Di}
\authornotemark[1]
\affiliation{
  \institution{AMAP, Alibaba Group}
  \city{Beijing}
  \country{China}
}
\email{difanyi.dfy@alibaba-inc.com}

\author{Yue Liu}
\affiliation{
  \institution{AMAP, Alibaba Group}
  \city{Beijing}
  \country{China}
}
\email{ly355576@alibaba-inc.com}

\author{Yifang Yuan}
\affiliation{
  \institution{AMAP, Alibaba Group}
  \city{Beijing}
  \country{China}
}
\email{yuanyifang.yyf@alibaba-inc.com}

\author{Jie Huang}
\affiliation{
  \institution{AMAP, Alibaba Group}
  \city{Beijing}
  \country{China}
}
\email{jielu.hj@alibaba-inc.com}

\author{Peng Wu}
\affiliation{
  \institution{AMAP, Alibaba Group}
  \city{Beijing}
  \country{China}
}
\email{zhouxiao.wp@alibaba-inc.com}

\author{Sicong Wang}
\affiliation{
  \institution{AMAP, Alibaba Group}
  \city{Beijing}
  \country{China}
}
\email{wsc488938@alibaba-inc.com}

\author{Mingyang Yin}
\affiliation{
  \institution{AMAP, Alibaba Group}
  \city{Beijing}
  \country{China}
}
\email{hengyang.ymy@alibaba-inc.com}

\author{Tingting Hu}
\affiliation{
  \institution{AMAP, Alibaba Group}
  \city{Beijing}
  \country{China}
}
\email{jingting.htt@alibaba-inc.com}

\author{Yao Xu}
\affiliation{
  \institution{AMAP, Alibaba Group}
  \city{Beijing}
  \country{China}
}
\email{xuenuo.xy@alibaba-inc.com}

\author{Xin Li}
\authornote{Corresponding Author.}
\affiliation{
  \institution{AMAP, Alibaba Group}
  \city{Beijing}
  \country{China}
}
\email{beilai.bl@alibaba-inc.com}

\renewcommand{\shortauthors}{Zhai et al.}

\begin{abstract}
The next point-of-interest (POI) recommendation task aims to predict the users’ immediate next destinations based on their preferences and historical check-ins, holding significant value in location-based services. Recently, large language models (LLMs) have shown great potential in recommender systems, which treat the next POI prediction in a generative manner. However, these LLMs, pretrained primarily on vast corpora of unstructured text, lack the native understanding of structured geographical entities and sequential mobility patterns required for next POI prediction tasks.
Moreover, in industrial-scale POI prediction applications, incorporating world knowledge and alignment of human cognition, such as seasons, weather conditions, holidays, and users' profiles (such as habits, occupation, and preferences), can enhance the user experience while improving recommendation performance. To address these issues, we propose CoAST (\textbf{Co}gnitive-\textbf{A}ligned \textbf{S}patial-\textbf{T}emporal LLMs), a framework employing natural language as an interface, allowing for the incorporation of world knowledge, spatio-temporal trajectory patterns, profiles, and situational information. Specifically, CoAST mainly comprises of 2 stages: (1) Recommendation Knowledge Acquisition through continued pretraining on the enriched spatial-temporal trajectory data of the desensitized users; (2) Cognitive Alignment to align cognitive judgments with human preferences using enriched training data through Supervised Fine-Tuning (SFT) and a subsequent Reinforcement Learning (RL) phase. Extensive offline experiments on various real-world datasets and online experiments deployed in "Guess Where You Go" of AMAP\footnote{https://amap.com/} App homepage demonstrate the effectiveness of CoAST.

\end{abstract}





\maketitle

\section{Introduction}
Next Point-of-Interest (POI) prediction, as a variant of POI recommendation task,
 aiming to predict users' future locations based on their historical sequential check-ins and other contextual information, is a fundamental task in Location-Based Services (LBS)
\cite{10.5555/2832415.2832536,9117156,chen2023stan,10.1145/3477495.3531983,10.1145/3539618.3591770,10.1145/3583780.3615083,10.1145/3637528.3671809,zhang2024llm4poi,wang2025generative,ye2018survey}.
For large-scale navigation platforms such as AMAP, which processes billions of daily requests from hundreds of millions of active users in China, making accurate next POI prediction a critical capability for efficient destination planning and personalized recommendation.
Thus, next POI prediction has been a popular research direction, capitalizing on developments in mobile and localization techniques, as they provide rich location-based information. 
Much of the existing research in next POI recommendation has predominantly focused on 
sequential recommendation and graph neural networks,
tackling the challenge of data sparsity, particularly for cold-start users and those with sparse historical trajectories \cite{10.1145/1772690.1772773, ye2011location, yuan2017pred, feng2018deepmove, Wang_2024}.


However, these approaches often fail to encapsulate the rich contextual information embedded within Location-Based Social Network (LBSN) data, such as the world knowledge and contextual information, which are critical for improving the user experience in industrial applications.
Recently, Large Language Models (LLMs) have demonstrated remarkable capabilities across various domains \cite{brown2020language, yang2025qwen3technicalreport, qwen2025qwen25technicalreport, touvron2023llamaopenefficientfoundation}, offering new opportunities for this direction.
However, their direct application to POI prediction task remains several key challenges. First, POIs are typically indexed by discrete identifiers outside the LLM vocabulary, impeding the model’s ability to capture semantic relationships \cite{Yu_2018, cao2024aligninglargelanguagemodels, 10597986}. 
Second, the intricate interactions between the billions of users and POIs, coupled with highly personalized and context-dependent behavioral patterns, introduce complex collaborative signals that are difficult for LLMs to encode and reason effectively \cite{liu2024nextlocllmlocationpredictionusing,chen2023stan}. 

Another critical challenge, stemming from the real-world deployment, is that recommended POIs must be cognitively plausible to users. Recommendations that violate human intuition or common-sense reasoning, which we term human cognitive preference in this work, even if statistically likely, can appear nonsensical or irrelevant, severely degrading the user experience and eroding trust in the system \cite{chen2023stan}. 
In industrial applications, these logical lapses are critical failures, as they directly risk frustrating users and reducing retention.

Based on these observations, we formulate the task of human cognitive preference alignment for individual POI prediction as: to investigate issues of personality and individual differences (\textbf{who}), environmental and social impacts (\textbf{what}), spatio-temporal features and relationships (\textbf{when} and \textbf{where}), and corresponding actions (\textbf{how}).

To address the aforementioned challenges and build a POI recommendation system that aligns with human cognitive preferences, we introduce CoAST (Cognitive-Aligned Spatio-Temporal LLM), a novel framework that endows a base LLM with the ability to generate cognitively plausible POI recommendations. CoAST begins by adapting the LLM to the mobility domain through continued pretraining on an enriched corpus of spatio-temporal tokenized POI sequences, allowing it to master complex user behavior patterns. Furthermore, we explicitly align the model with human common-sense. This is achieved through a multi-stage fine-tuning process, utilizing a curated dataset of cognitively-aware samples for both Supervised Fine-Tuning (SFT) and a subsequent Reinforcement Learning (RL) phase.


Our main contributions as as follows:
\begin{itemize}
    \item To our knowledge, this is the first generic LLM-based cognitive-aligned framework for the field of next POI prediction. The proposed framework enhances the comprehension of human cognitive preference in recommendation.
    \item We propose CoAST, a novel end-to-end LLM-based framework for large-scale next POI prediction that effectively bridges the semantic gap between world and recommendation knowledge, which demonstrates significant improvement on human cognitive alignment and applicable to practical applicability of LLMs in LBS applications.
    \item Through a series of inference optimization strategy, CoAST is deployed in AMAP's online production environment, a leading real-world LBS platform. Compared traditional cascade ranking system,  
    long term online A/B test achieves a 4.23\% and 4.59\% increase in P-CTR and U-CTR, which is a substantial improvement. 
\end{itemize}

The remainder of this paper is organized as follows: Section \ref{sec:related} reviews related literature, Section \ref{sec:prelim} details problem definition and the detailed categories of cognitive recommendation, Section \ref{sec:method} elaborates our methodology, Section \ref{sec:exp} reports empirical results, and Section \ref{sec:conclusion} concludes the paper.

\section{Related Work}
\label{sec:related}
\subsection{Next POI Prediction}
Next POI recommendation extends beyond conventional item recommendation by incorporating inherent geographical and temporal characteristics. Early methods relied on user check-in histories using Markov chains, or RNNs to capture sequential behaviors. 
Spatial-temporal modeling has been extensively studied with works like ST-GRAT \cite{Park_2020} employing graph attention networks for spatial-temporal dependencies, STAN \cite{chen2023stan} introducing hierarchical attention mechanisms, and Sun et al. \cite{sun2021periodicmove} focusing on periodic patterns in human mobility through time-aware embeddings.
Recent LLM-based approaches have focused on addressing the semantic gap in POI recommendation. For example, 
GNPR-SID \cite{wang2025generative} proposed semantic POI IDs using residual quantized variational autoencoders, achieving improvements by mapping POIs into discrete semantic spaces. LLM4POI \cite{zhang2024llm4poi} leveraged pretrained LLMs to preserve heterogeneous data in original format, effectively utilizing contextual information. 

\subsection{LLMs for Recommendation}
The integration of Large Language Models (LLMs) into recommendation systems has emerged as a transformative research direction. P5 \cite{geng2023recommendationlanguageprocessingrlp} pioneered the "Pretrain, Personalized Prompt, and Predict" paradigm, establishing the foundation for treating recommendation as a language processing task by converting all recommendation data into natural language sequences. Building upon this, \cite{cao2024aligninglargelanguagemodels} addressed the knowledge gap between LLMs' capabilities and recommendation-specific requirements by incorporating collaborative filtering operations like Masked Item Modeling (MIM) and Bayesian Personalized Ranking (BPR) through natural language simulation. More recently, GenRank \cite{huang2025largescalegenerativeranking} advanced generative ranking systems for large-scale deployment. OneRec \cite{deng2025onerecunifyingretrieverank, zhou2025onerectechnicalreport} proposed a revolutionary end-to-end framework that replaces traditional cascaded architectures, achieving significant improvements in large-scale online deployments.

\section{Preliminaries}
\label{sec:prelim}
\subsection{Problem Formulation}
Given a collection of POIs $\mathcal{I}=\{p_1, p_2,...,p_N\}$ and a group of users $\mathcal{U}=\{u_1, u_2,...,u_M\}$. Each POI can be characterized by a tuple $p_i = (lon,lat,c)$,  where $c$ represents side information of the POI including the descriptive texts and pictures, $lon$ and $lat$ denotes the geographical coordinates (longitude and latitude). 
Each user $u$ has the check-ins history $h_u = \{(p_1, t_1), \ldots, (p_m, t_m)\}$ of length $m$, where $p_i, i = 1, \ldots, m$ reflects the $i$-th POI recently interacted by the user and $ t_i$ refers to the check-in timestamps.
The objective of next POI prediction task is to predict the next POI $p$ that the user $u$
will visit at time $t$ given the recent check-in record $h_u$, the user's request location $l$ and the contextual situation $s$.

\subsection{Cognitive-Aligned POI Prediction}
\label{sec:cognitive}
Compared to traditional accuracy-oriented recommender systems that rely on rapidly capturing users' immediate interests,  next POI prediction is a complex issue in simulating individual behaviors and contextual information.
Empirically, we category cognitive-aligned recommendations into the following four scenarios. Recommended POIs that do not align with these categories may result in a diminished user experience or decreased user engagement.
To summarize, our approach formulates the next POI recommendation task as a decision-making problem. The model synthesizes a diverse set of contextual factors—including temporal (when), spatial (where), user-specific (who), and other situational inputs—to determine the optimal user action (what to do and how).

\subsubsection{Temporal Consistency} User behavior often exhibits certain regularities. For example, office workers typically commute to workplaces in the morning and return home at noon. 
Recommending irrelevant POIs during these commuting periods may lead to poor user experiences.
We define these observed regularity patterns in user mobility as temporal consistency. Formally, at the given time $t$, the temporal consistency score (denoted as TCS) of the predicted POI $\hat{p}$ can be formulated as,
\begin{equation}
\text{TCS}(u, l, s, t,\hat{p}) = \frac{\sum_{i=m}\mathbb{I}(t_i=t) \cdot \text{sim}(p_i, \hat{p}*)}{\sum_{i=1}^{m} \mathbb{I}(t_i=t)}
\end{equation}
where $\text{sim}(\cdot)$ denotes the function measuring the similarity of the inputs pair.

\subsubsection{Spatial Clustering}
Check-ins of human are empirically demonstrated to occur within a certain range \cite{10.1145/2623330.2623638,10.1145/3182166}; that is, the distance of a user's next visit POI is typically distributed as a negative exponential relative to the user's usual activity space.
Recommending a POI that is far from the user's position (or their active area) is inappropriate. 
We define the spatial clustering phenomenon of user mobility as spatial consistency. Formally, given the location $p_u$ of the user $u$,
the spatial consistency score of the predicted POI $\hat{p}$ is formulated as,
\begin{equation}
\text{SCS}(u, l, s, t, \hat{p}) = e^{-{d\left(p_u, \hat{p} \right)}/{d_u}}
\end{equation}
where $e$ denotes the exponential function, $d(\cdot)$ is the distance function, and $d_u$
represents the user’s historical average movement distance.

\subsubsection{Profile Alignment}
Behaviors of human often correlate with their profile tags. For instance, suggesting family-oriented venues or restaurants to university students represents a cognitive mismatch in recommendation logic.
We define the user profile alignment score to reflect the alignment between recommended item $\hat{p}$ and user characteristics:
\begin{equation}
\text{PAS}(u, l, s, t, \hat{p}) = \text{LLM}_{u}(u, \hat{p})
\end{equation}
where a LLM model is utilized to evaluate the reasonableness of the recommended POI $\hat{p}$ given the profile of user $u$ and the score $\in \{0, 1\}$.

\begin{figure*}
    \centering
    \includegraphics[width=\linewidth]{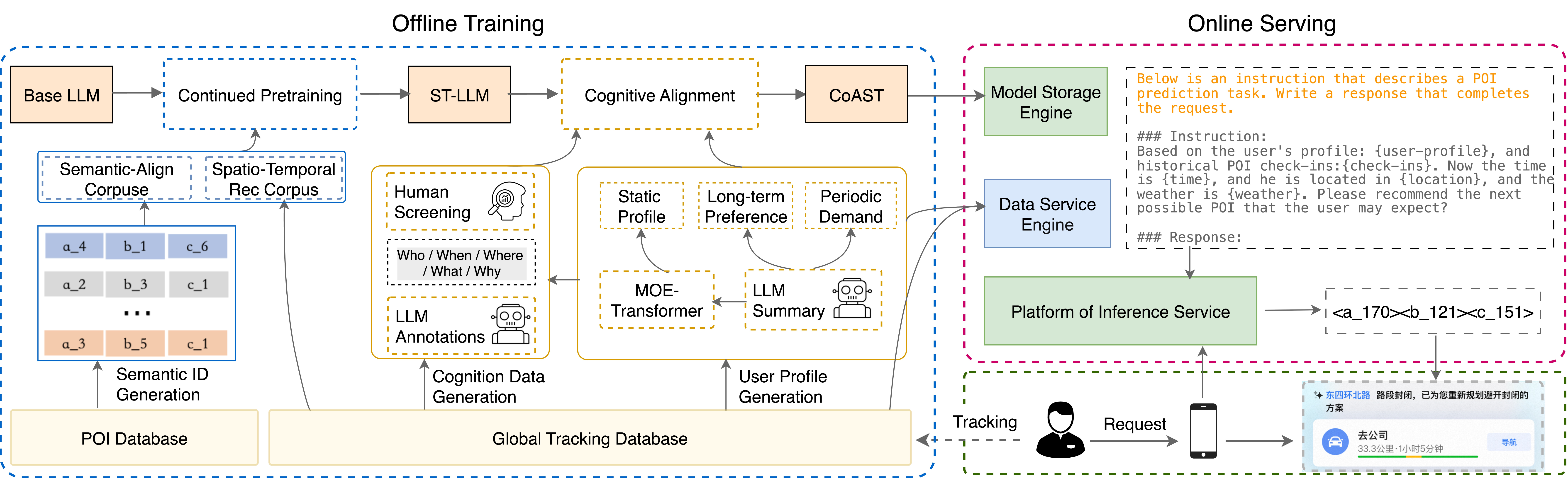}
    \caption{Offline Training and Online Service Procedure of proposed method CoAST. Offline Training: (1) POIs are tokenized as Semantic IDs (SIDs); (2) a base LLM is continually pretrained on the collected corpus, yielding ST-LLM; (3) multi-level user profiles are distilled from the global tracking database with the annotations of LLM; (4) cognitive-aligned data are produced through LLM annotation and human screening; (6) ST-LLM is further fine-tuned and aligned to finish this procedure. Online Serving: CoAST is stored in a model engine and invoked through a data-service engine that converts real-time context—user profile, recent check-ins, current time, location, and weather—into an instruction prompt. The model predicts the next POI in SID form, which is mapped back to a readable location and returned to the mobile client. }
    \label{fig:framework}
\end{figure*}

\subsubsection{Situational Awareness}
Situations such as weather, traffic, and the conditions of the POIs could have a significant impact on human's traveling decisions. For example, recommending an outdoor tennis court on a rainy day is considered inappropriate. We define the recommendation system’s Awareness to these contextual situations as situational awareness. Similar to definition of profile alignment, we also utilize a LLM to score this metric,
\begin{equation}
\text{SAS}(u, l, s, t, \hat{p}) = \text{LLM}_{s}(s, \hat{p})
\end{equation}
where the score $\in \{0, 1\}$.

The definitions articulated above serve a dual purpose in our methodology. They directly guide the creation of a training set of cognitively plausible examples for the model's alignment phase. Simultaneously, these definitions are translated into quantitative metrics, providing a principled framework for evaluating the cognitive capabilities of our proposed model.


\section{Methodologies}
\label{sec:method}
The overall framework is presented in Figure \ref{fig:framework}. The framework consists of two main modules: (1) Spatio-temporal Knowledge Acquisition module enables the LLMs to incorporate the linguistic and collaborative semantics and (2) Cognitive Alignment to adjusting the ST-LLM obtained from step (1) to generate results that better align with human cognition. We finally detail our inference optimization strategy in online deployment since this is a bottleneck of large-scale application of generative models.

\subsection{Spatio-temporal Knowledge Acquisition}
Considering the huge amount of the POI corpus in large industrial LBS applications, we firstly convert POI data into Semantic IDs (SIDs) to reduce the total number of tokens. Building on these SIDs, we propose to enable the LLM to incorporate the recommendation knowledge by continued pretraining on an enriched corpus.

\subsubsection{Semantic ID generation}
A well crafted SID representation can boost the LLM's recommendation efficiency by capturing the semantic and spatial relationships among POIs. There has already been extensive research in this area \cite{zheng2025universalitemtokenizationtransferable,zhai2025simcit,10.1145/3640457.3688178,10.1145/3627673.3679569,wang2025generative,10.1145/3726302.3729989}, so we do not elaborate on it here. In brief, we adopt the RQ-VAE, which adopts the multi-modal information of the POIs to learning the codebooks.
Specifically, we incorporate the spatial location information and descriptive titles/pictures of the POIs to generate the identifiers with residual
quantization process. As a result, we get a set of codebooks
\begin{equation*}
    C_l = \{\boldsymbol{e}_k^l|k = 1, ..., K\},  l \in [1,L] 
\end{equation*}
where $L$ represents the number of codebooks and $K$ denotes the size of each codebook.  Therefore, a POI can be tokenized as an ordered sequence of tokens, such as "<a\_17> <b\_21><c\_119>", in which the number of the above tokens represents the indices of the responding codebook.

\subsubsection{Continued Pretraining on Spatio-temporal Corpus}
After acquiring SIDs of the POIs, a simple method is to incorporate these SIDs into the vocabulary of large language models (LLMs). This allows the LLMs to perform recommendation tasks in a generative manner, ultimately moving away from using item indices. 
Nevertheless, there are two remaining concerns: 
(1) these newly added indices are out-of-vocabulary tokens for LLMs, incorporating these linguistic and collaborative semantics becomes necessary. (2) Recommendation tasks present unique technical challenges that distinguish them from conventional NLP tasks, particularly in modeling spatio-temporal user behaviors and handling data sparsity.

To achieve this incorporation, we implemented continued pretraining \cite{gururangan-etal-2020-dont} on the recommendation corpus, aiming to adjust LLMs to improve understanding of linguistic and collaborative semantics.
Our approach to recommendation domain-adaptive pretraining is straightforward---we continue pretraining a base model (like Qwen3 \cite{yang2025qwen3technicalreport}) on a large corpus of unlabeled spatio-temporal recommendation domain-specific text.
Specifically, the corpus are designed and collected mainly
focusing on 2 objectives:
clear index-language-space aligning (finding the POI with its description, location and its category, or outputting the description and location given the POI) and sequential POI prediction (predicting the next POI given the historical POI check-ins).

While understanding the spatial relationships between POIs is fundamental to LBS recommendation, directly instilling this knowledge into LLMs is difficult. To address this, our approach tokenizes each POI's Geohash and integrates these tokens into our pretraining data. This process enables the model to learn a unified embedding space where the POI's identity (SID), its semantic meaning, and its location (via Geohash) are jointly represented and aligned.
Table \ref{tab:pretrain-corpus} illustrates the utilized corpus in this stage. 

\begin{table}[h]
\centering
\caption{Recommendation corpus collected for continuous pretraining based on a base model. The words in orange denote the newly added tokens in the base LLM.}
\label{tab:pretrain-corpus}
\begin{tabular}{p{8cm}}
\toprule[1pt]
\textbf{SID-Location-Description Alignment Corpus} \\ 
\midrule
The Temple of Heaven Park is the world’s largest surviving ancient architectural complex dedicated to the worship of heaven. Originally constructed during the Ming Dynasty, it served as the site where emperors of the Ming and Qing dynasties held ceremonies to “sacrifice to heaven” and “pray for a good harvest.” Located in \textcolor{orange}{<wm6j0>} of Beijing, and the ID is \textcolor{orange}{<a\_{82}><b\_{59}><c\_{191}>}. 
\\ 
\midrule
\textbf{Spatio-Temporal Behavior Sequence Recommendation Corpus} \\ 
\midrule
Here is user uid\_256’s historical POI check-ins. At 7 AM June 10, the user ride to company <a\_124><b\_{192}><c\_{41}>, at 21 PM June 11, the user \textcolor{blue}{navigated to} \textcolor{red}{the hotel} \textcolor{orange}{<a\_{82}><b\_{59}><c\_{191}>}, at 15 PM June 20, the user \textcolor{blue}{searched} \textcolor{red}{an office} \textcolor{orange}{<a\_12><b\_{28}><c\_{140}>},..., at 19 PM July 20, the user \textcolor{blue}{walked to} \textcolor{red}{home} \textcolor{orange}{<a\_124><b\_{192}><c\_{41}>}.
\\
\bottomrule[1pt]
\end{tabular}
\end{table}

Masked prediction has recently been applied to adapt pretrained models to specific downstream
tasks by continuing to pretrain models on in-domain unlabelled data \cite{wilf2023differencemaskingchoosingmaskcontinued, dery2023aangautomatingauxiliarylearning, 5ae57eb26ea74cf28cc864d52301e6fd}. 
Inspired by these works, we continue our pretraining with masking some original tokens of the base LLM.
However, a crucial consideration in our pretraining phase is the token composition of the training corpus, which presents a delicate trade-off
between newly introduced POI tokens (SIDs and Geohash tokens) and the original tokens within the pretraining corpus. 


Through the comprehensive continued pretraining on the corpus of spatio-temporal recommendation domain, the model is adapted to learn information centered around generic behavioral patterns.
However, its outputs may exhibit cognitive mis-alignment with human reasoning principles.
This necessitates a specialized cognitive calibration process to bridge the divergent representations between machine-learned patterns and anthropic decision-making frameworks.

\subsection{Cognitive Preference Alignment}
\subsubsection{User Profile Generation}
Understanding user historical POI check-ins is crucial for modeling the trajectory patterns of users. However, it is challenging to effectively utilize extended user behavior sequences, which often extends to lengths of 10,000 in large-scale recommendation systems \cite{10.1145/3589335.3648334, 10.1145/3637528.3671601}.
Yet, LLMs face difficulties in processing these long sequences, and this hinders their ability to grasp the patterns.
To counteract this, we have developed multi-level profile of the users
by condensing extensive user behavior into refined profiles that are more amenable to LLMs. 
The utilized profiles can be mainly divided into three main categories: (1) Static Profile, (2) Long-term Preference Profile, and (3) Periodic Demand Profile. 

Static profile, such as POIs of the users' home and company, are critical for next POI prediction. However, the labeled data is pretty limited. 
To expand the coverage of static profiles, we first employ LLM-based summarization of user behavioral sequences to generate preliminary annotations. These labeled patterns are then utilized to train models (like a MOE-Transformer network \cite{10.5555/3586589.3586709}) that subsequently perform inference across the entire user population, producing comprehensive static profiles that maintain semantic consistency with human cognitive patterns.

Long-term preference profile refers to stable and long-term intentions derived from long-term behaviors, which focuses more on stability, accuracy, and long-term relevance, requiring a more rigorous reasoning and calibration process \cite{xi2025burstingfilterbubbleenhancing, yi2025recgpttechnicalreport}.

Periodic demand profile identifies recurring consumption patterns that exhibit temporal periodicity correlated with interval cycles (e.g., weekly/monthly), holiday events, and workday rhythms. 
We employs a LLM to summarize the extended behavioral sequences through scheduled interval sampling, implementing an automated periodicity and long-term preference
detection mechanism.


\subsubsection{Cognitive-aligned Data Generation}
Within the vast amount of POI check-in data left by users, a significant portion is considered noise (such as accidental clicks), or does not align with the human cognition process.
Therefore, we intervene in data collection by combining the world knowledge of
LLMs and human annotations. Following similar process of \cite{xi2025burstingfilterbubbleenhancing}, we collect the data from traditional CTR-oriented recommender systems that align with human cognition preference defined in Section \ref{sec:prelim}.

In particular, LLMs is utilized to access the users’ historical check-ins, profile information and other contextual information (like time and weather)
to conduct chain-of-thought (CoT) reasoning \cite{10.5555/3600270.3602070}, thereby evaluating the potential candidate items aligned with human cognition for individual users. 
Subsequently, human annotators review the reasoning process and outcomes generated by the LLMs to refine and improve the accuracy of serendipity labels.
This annotated dataset—which contains user history, profiles, candidate items,
contextual conditions,
and cognition annotations—is then utilized to further fine-tune the model after continued pretraining phase that can identify the examples aligned with human cognition for the subsequent training. This approach enables scalable discovery of cognitive-aligned  items while reducing the need for extensive manual annotation.

\subsubsection{Supervised Fine-tuning}
After obtaining cognitive-aligned data, with each record consisting of user $u$, 
their multi-level profiles, historical POI check-ins $h_u$, the current situation of the user (time, location, weather and so on)
and the selected cognitive-aligned POI $p_i$, we begin the process of fine-tuning by aligning the inputs.
Initially, all of the inputs are transformed into a text-based supervised fine-tuning (SFT) dataset, where each record is a pair $(x, y)$, with $y$ denoting the output SID of the target POI $p_i$, and $x$ as the input prompt. For example, 
\begin{tcolorbox}
\textbf{Instruction:} \\
Based on the user's profile: \{\textcolor{red}{user-profile}\}, and his historical POI check-ins: \{\textcolor{red}{check-ins}\}. Now the time is \{\textcolor{red}{time}\}, he is in \{\textcolor{red}{user-location}\}, and the weather is \{\textcolor{red}{weather}\}. Please recommend the next possible POI that the user may expect?\\
\textbf{Response:} \\
<a\_{112}><b\_{32}><c\_{5}>
\end{tcolorbox}
Subsequently, LLM is
fine-tuned under supervision to enhance its ability to produce the desired responses.

\subsubsection{Cognitive Preference Alignment}
Following the standard paradigm for human preference alignment, we continue to leverage Reinforcement Learning (RL) to further refine the model's outputs and explicitly align its recommendations with human cognitive preferences. 

After obtaining datasets comprising human-annotated preferences over pairs of model-generated responses, we train the model with the preference alignment algorithm.
We choose Direct Preference Optimization (DPO) \cite{10.5555/3666122.3668460} in this phase, since it has been widely employed to align the model’s outputs more closely with human preference.
\begin{equation}
\mathcal{L} = -\mathbb{E}_{(x, y_w, y_l) \sim D} \left[  \log \sigma \left( \beta\frac{\pi_{\theta}(y_w \mid x) }{\pi_{\text{ref}}(y_w \mid x) }  -\beta \frac{\pi_{\theta}(y_l \mid x)}{\pi_{\text{ref}}(y_l \mid x)}\right) \right]
\end{equation}
where \( D \) denote the preference dataset, \( x \) denotes the input prompt, and \( y_w \) and \( y_l \) represent the preferred and dis-preferred responses annotated by human and LLM. \( \pi_{\theta} \) is the trainable LLM for preferences alignment, while \( \pi_{\text{ref}} \) is a fixed reference policy. \( \tau \) is the hyper-parameter of the regularization term balancing learning human preferences and staying close to the reference policy, thereby preventing overfitting. 


In summary, after the Supervised Fine-Tuning (SFT) phase, our model adheres to the required output format (a list of $L$ valid SIDs) and already possesses a baseline level of cognitive awareness from the training data. The subsequent DPO phase then significantly enhances this cognitive reasoning while preserving the model's recommendation performance.


\subsection{Online Deployment}
A primary challenge for the large-scale application of generative models is their inference latency, particularly in recommendation scenarios with high Queries Per Second (QPS).
To enable the large-scale online deployment of our generative model, 
inspired by recent works \cite{10.5555/3692070.3692699, 10.5555/3691938.3691949}, we implemented a dual-pronged optimization strategy targeting both system architecture and algorithmic efficiency, achieving significant breakthroughs in stability and performance.

\subsubsection{Prefill-Decoding Decoupling}
We employ Prefill-Decoding (PD) Decoupling \cite{10.5555/3691938.3691949} to separate the inference process into two distinct stages: a computationally-intensive Prefill stage for prompt processing and a memory-bound Decoding stage for iterative token generation. By applying dedicated optimization strategies to each stage, this approach mitigates resource contention and latency jitter common in unified pipelines. 
The resulting improvements in resource scheduling and concurrency led to a 10-fold increase in throughput, drastically reducing the per-inference cost on the same hardware.
\subsubsection{Multi-Token Prediction}
Furthermore, we leverage Multi-Token Prediction (MTP) to overcome the sequential dependencies of traditional auto-regressive generation \cite{10.5555/3692070.3692699}. Specifically, we adopt MTP Eagle, an optimized variant of the Eagle speculative decoding method. Its key innovation lies in a single, reusable MTP module that shares the same KV cache for predicting multiple draft tokens, thus eliminating the redundant computational overhead of vanilla MTP. Its chained decoding strategy, which only requires the last predicted token and its hidden state to proceed, significantly simplifies implementation by obviating the need to store historical states. MTP Eagle is fully compatible with existing checkpoints (e.g., DeepSeek-V3) and substantially reduces both memory footprint and computational complexity, enhancing the engineering feasibility and resource efficiency of our deployment.

Collectively, these optimizations create a highly efficient and stable inference framework. This is crucial for deploying large-scale generative models in production, ensuring both a responsive user experience and the economic viability of the system in a real-world advertising setting.

\section{Experiments}
\label{sec:exp}

\subsection{Experimental Settings}

\subsubsection{Scenario} Our method is performed on \textbf{AMAP}, a prominent navigation and mapping platform hosting billions of users and POIs. We focus on the "\textbf{Guess Where You Go}" card presented on the homepage of the AMAP App, aiming to forecast the next subsequent POI using a user’s previous check-ins and profile information. 
The objective of this scenario is to predict the next POI the users are most likely to visit. By displaying the POIs on the homepage, we aim to reduce the users' search effort, thereby enhancing user experience and long-term retention. Critically, our goal transcends mere prediction accuracy; we strive to build a model that captures the underlying dynamics of user mobility patterns and intents. This allows the system to develop a deeper "cognitive" understanding, ultimately facilitating more intelligent and context-aware decision-making for the user.

\begin{figure}[h!]
    \centering
    \includegraphics[width=\linewidth]{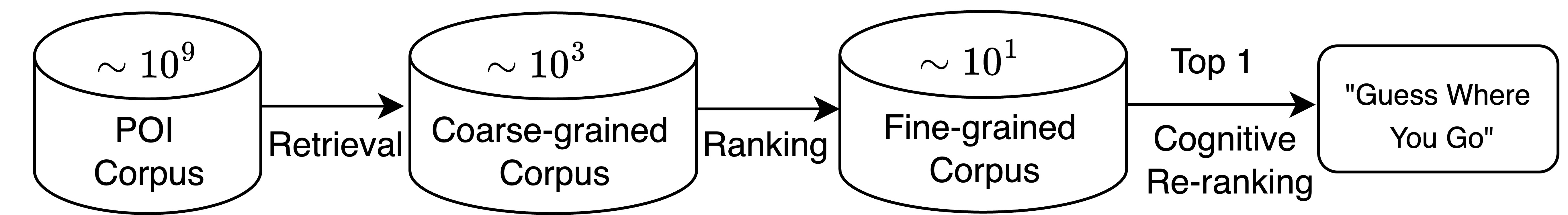}
    \caption{Cascade ranking system of "Guess Where You Go" deployed on AMAP, which includes three stages from the left to the right: Retrieval, Ranking, and Cognitive Re-ranking.}
    \label{fig:cascade}
\end{figure}

The current deployed framework, which serves as our primary baseline, employs a traditional cascaded ranking system for recommendation. As illustrated in Figure \ref{fig:cascade}, the cascaded ranking system includes three stages: (1) Retrieval stage to recall a coarse-grained POI corpus based on the user's preference, historical check-ins, and location \cite{10.1145/3357384.3357814,yang2020large}; (2) Ranking stage to obtain top 10 POIs the users most likely visit with a ranking model based on behavior sequence transformer \cite{10.1145/3326937.3341261}; and (3) Cognitive Re-ranking stage with distance, quality, and exposure filters to yield the most cognitively plausible destination.

{\color{blue}\subsubsection{Datasets}}
We evaluate our approach on three real-world datasets: Foursquare-NYC \cite{6844862}, Foursquare-TKY \cite{6844862}, and Gowalla-CA \cite{10.1145/2020408.2020579} and a large-scale proprietary dataset by sampling desensitized users from AMAP (named Beijing). The details of the utilized dataset is illustrated in Appendix \ref{sec:dataset}.

\subsubsection{Implementation Details}
The quantization module consists of 3 codebook layers and 
for the generative recommendation module, we employ the Qwen \cite{qwen2025qwen25technicalreport, yang2025qwen3technicalreport} as the base model. We employ a constant learning rate schedule with a learning rate of 1e-5, combined with a warm-up phase of 20 steps. The model is trained on 200$\times$ NVIDIA H20 GPUs, with a batch size of 16 per GPU, gradient accumulation steps set to 8, and a sequence length of 4096 tokens. Each input consists of the most recent 50 check-in records of the user.

\subsubsection{Evaluation Metric}
Following previous works, we employ the accuracy metric for the top-1 recommendation accuracy (\text{Acc}@1) and the four cognitive metrics (TCS, SCS, PAS and SAS) defined in \ref{sec:cognitive} as our metrics to evaluate the effect of the cognitive recommendation. Detailed information of the metrics is illustrated in Appendix \ref{sec:evaluation}.

\begin{table}[h!]
\centering
\caption{Comparison of different models on three datasets: NYC, TKY, and CA. We present the model inputs including the POI and User representation approach and whether visit timestamps are utilized. RID, SID, UID and UP represent Random one-hot ID, Semantic ID, User one-hot ID and User Profiles, respectively.  The results for the baseline methods are borrowed from \cite{zhang2024llm4poi, wang2025generative}.}
\label{tab:nyc}
\setlength{\tabcolsep}{4.pt}
\begin{tabular}{llcccccc}
\toprule
\multirow{2}{*}{Model} &  \multicolumn{3}{c}{Inputs} & & \multicolumn{3}{c}{Acc@1} \\
        \cmidrule{2-4} \cmidrule{6-8}
      & POI & User & Time  & & NYC & TKY & CA \\
\midrule
PRME    & RID & UID & $\times$ & & 0.1159 & 0.1052 & 0.0521 \\
PLSPL   & RID & UID & $\times$ & & 0.1917 & 0.1889 & 0.1072 \\
STAN    & RID  & UID & $\times$ & & 0.2231 & 0.1963 & 0.1104 \\
GETNext & RID & UID & $\times$ & & 0.2435 & 0.1829 & 0.1357 \\
STHGCN  & RID & UID & $\times$ & & 0.2734 & 0.2950 & 0.1730 \\
TPG     & RID  & UID & $\checkmark$ & & 0.2555 & 0.1420 & 0.1749 \\
ROTAN   & RID  & UID & $\checkmark$ & & 0.3106 & 0.2458 & 0.2199 \\
LLM4POI & RID & UID & $\checkmark$ & & 0.3372 & 0.3035 & 0.2065 \\
GNPR-SID & SID & UID & $\checkmark$ & & 0.3618 & 0.3062 & 0.2403 \\
\midrule
CoAST-0.5B & SID & UP & $\checkmark$ & & 0.3705 & 0.3125 & 0.2555 \\
CoAST-1.5B & SID & UP & $\checkmark$ & & 0.3727 & 0.3170 & 0.2585 \\
CoAST-3B & SID & UP & $\checkmark$ & & 0.3872 & 0.3202 & 0.2611 \\
CoAST-7B & SID & UP & $\checkmark$ & & \textbf{0.4027} & \textbf{0.3310} & \textbf{0.2721} \\
\bottomrule
\end{tabular}
\end{table}

\subsubsection{Baseline Methods}
The details of the baseline methods in this part is illustrated in Appendix \ref{sec:baseline}.

\subsubsection{Our Models} In our experiments we consider three versions of our model: 
(i) CoAST-0.5B: the model trained with Qwen2.5-0.5B \cite{qwen2025qwen25technicalreport} as our base LLM. 
(ii) CoAST-1.5B: with Qwen2.5-1.5B as our base LLM. (iii) CoAST-3B: A variation with Qwen2.5-3B as our base LLM. (iv) CoAST-7B: A second variation on Qwen2.5-7B. Below, unless stated otherwise, when we say “our model,” we refer to the CoAST-0.5B.

\begin{table}
  \caption{Comparison results of different versions of CoAST on Beijing dataset with the Cascade Ranking baseline system. 'a-' denotes average.}
  \label{tab:beijing}
  \setlength{\tabcolsep}{3.5pt}
  \begin{tabular}{cccccc}
    \toprule
    &\small Baseline &\small CoAST-0.5B& \small CoAST-1.5B&\small CoAST-3B&\small CoAST-7B\\
    \cmidrule{2-6}
    Acc@1 & 0.2675 & 0.2820 & 0.2862  & 0.2923 & 0.3071 \\
    a-TCS & 0.5561 & 0.6071 & 0.6021 & 0.6632 & 0.7543 \\
    a-SCS & 0.8022 & 0.8212 & 0.8231 & 0.8452 & 0.8838 \\
    a-PAS & 0.6232 & 0.6285 & 0.6240 & 0.6531 & 0.6970 \\
    a-SAS & 0.2472 & 0.2840 & 0.2988 & 0.3240 & 0.3725 \\
  \bottomrule
\end{tabular}
\end{table}

\subsection{Offline Experiments}
\subsubsection{Overall Performance}
Our method substantially outperforms all baseline methods across all of the evaluated datasets. Table \ref{tab:nyc} illustrate the comparison results on NYC, TKY, and CA datasets. Specifically, compared to the state-of-the-art GSNR-SID which utilizes LLaMA-7B as its base model, we observe a substantial improvement on all of the versions of CoAST. Notably, CoAST-7B demonstrated the most significant improvements, achieving gains of 11.3\%, 8.1\%, and 13.2\% over the SOTA, respectively. 

Table \ref{tab:beijing} details the performance comparison between our proposed model and the cascade ranking system on the Beijing dataset. In addition to the standard Acc@1 metric, we evaluate the models on four cognitive-aware recommendation metrics (where 'a-' denotes average).
Consistent with our findings on other datasets, our model significantly outperforms the traditional cascade ranking system on Acc@1. More critically, it demonstrates substantial gains on the cognitive metrics, achieving remarkable uplifts of 35.6\% in temporal consistence and 50.68\% in situational awareness performance. These results highlight a key limitation of conventional collaborative filtering-based systems: their struggle to capture personalized, periodic user needs and inability to effectively model situational or contextual information. This validates the superiority of our generative approach, which excels in understanding these complex, dynamic aspects of user behavior.




\subsubsection{Ablation Study}
To evaluate the impact of individual modules, we developed various versions of CoAST. Initially, certain essential modules were removed: "w/o UP" eliminates user profiling from the prompt. "w/o S-ST" eliminates situational and spatio-temporal descriptions from the prompt.
"w/o SLD" exclude the SID-Location-Description alignment corpus in the continued pretraining stage.
"w/o CPT" exclude the Continued Pretraining stage in the alignment stage, only applying SFT and DPO.
As illustrated in Table \ref{tab:ablation}, removing any module leads to noticeable performance degradation, emphasizing the importance of each component. Significantly, the lack of SFT results in the most considerable decrease in all metrics, demonstrating the essential role of preference alignment in recommendation efficiency. 

\begin{table}[h!]
    \centering
    \caption{Ablation study of CoAST (with the online deployment version CoAST-3B).}
    \label{tab:ablation}
    \begin{tabular}{lccccc}
        \toprule
         & Acc@1 & a-TCS &  a-SCS & a-PAS & a-SAS  \\
        \midrule
        w/o UP & 0.2720 & 0.6435 & 0.8211 & 0.5920 & 0.2923  \\
        w/o S-ST & 0.2708 & 0.5905 & 0.7938 & 0.6496 & 0.2822  \\
        w/o SLD & 0.2825 & 0.6078 & 0.8240 & 0.6501 & 0.3020  \\
        w/o CPT & 0.2632 & 0.5875 & 0.7820 & 0.6020 & 0.2875  \\
        \midrule
        CoAST & 0.2923 & 0.6632 & 0.8452 & 0.6531 & 0.3040  \\
        \bottomrule
    \end{tabular}
\end{table}

\subsubsection{Case Study}
To provide a qualitative and intuitive demonstration of our model's advantages, we conduct a case study that contrasts the recommendation lists generated by our model with those from a baseline model for a representative user. Detailed information of the case is illustrated in Appendix \ref{sec:casestudy}.

\subsubsection{Further Analysis}
\begin{figure}[h!]
    \centering
    \includegraphics[width=\linewidth]{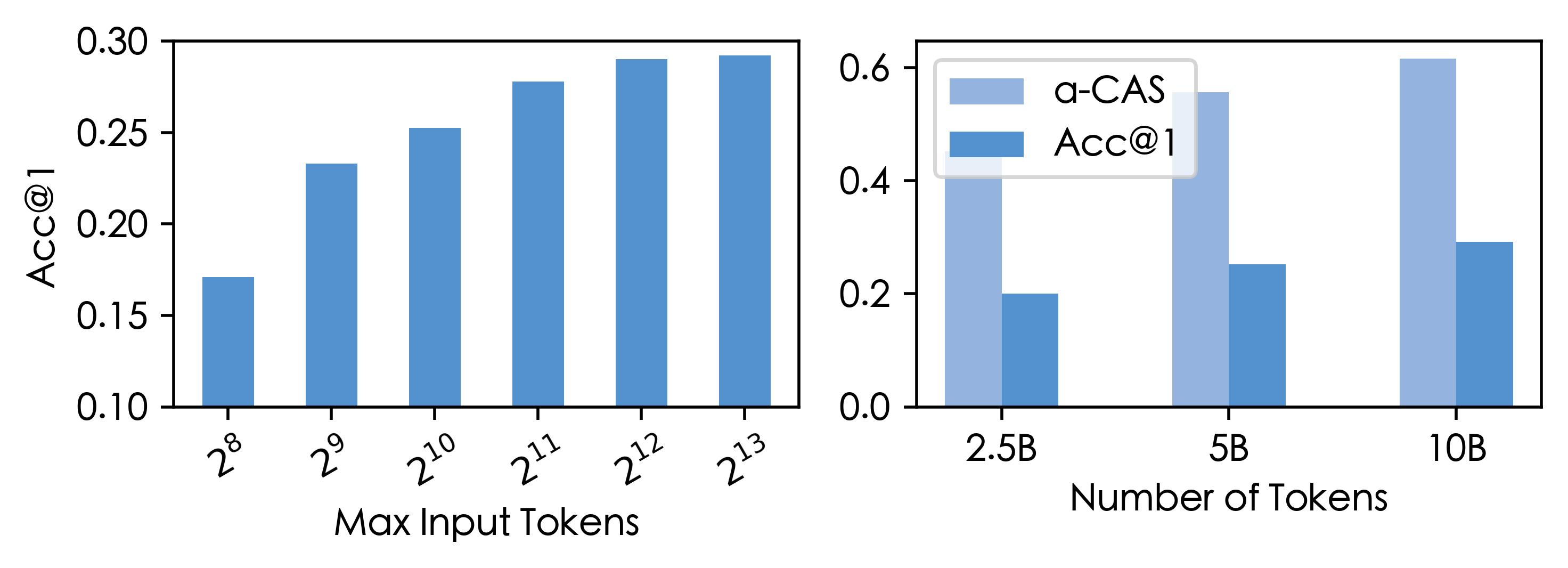}
    \caption{Impact on the performance of sequence length (input tokens) and number of tokens involved in continued pretraining stage. a-CAS denotes the average Cognition Alignment score computed by averaging the four cognitive scores.}
    \label{fig:impact}
\end{figure}

In this section, we analyze the impact on the performance of sequence length (input tokens) and number of utilized tokens in continued pretraining stage. Figure \ref{fig:impact} presents the results.
Our investigation into the impact of input sequence length shows that while longer context is beneficial, its utility eventually saturates. The model's performance gains are most pronounced up to a context window of 2048 tokens ($\sim$30 user historical check-ins) and largely stabilize beyond 4096 tokens ($\sim$50 check-ins). This suggests that a moderately-sized history is sufficient to encapsulate the user's core preferences. Further extending the context window provides limited new signal, as this long-term information is already implicitly encoded in the user's learned representation.

The volume of tokens involved in pretraining phase also proved to be a key determinant of performance. We observed a nearly linear improvement in model effectiveness as the training corpus size increased. Constrained by our training budget, the final model was trained to convergence on a dataset of 10B tokens. The consistent, non-saturating growth suggests that performance is largely data-bound, with significant potential for improvement as more data becomes available.

\subsection{Online Experiments}

\subsubsection{Efficiency Analysis}
\begin{figure}
    \centering
    \includegraphics[width=\linewidth]{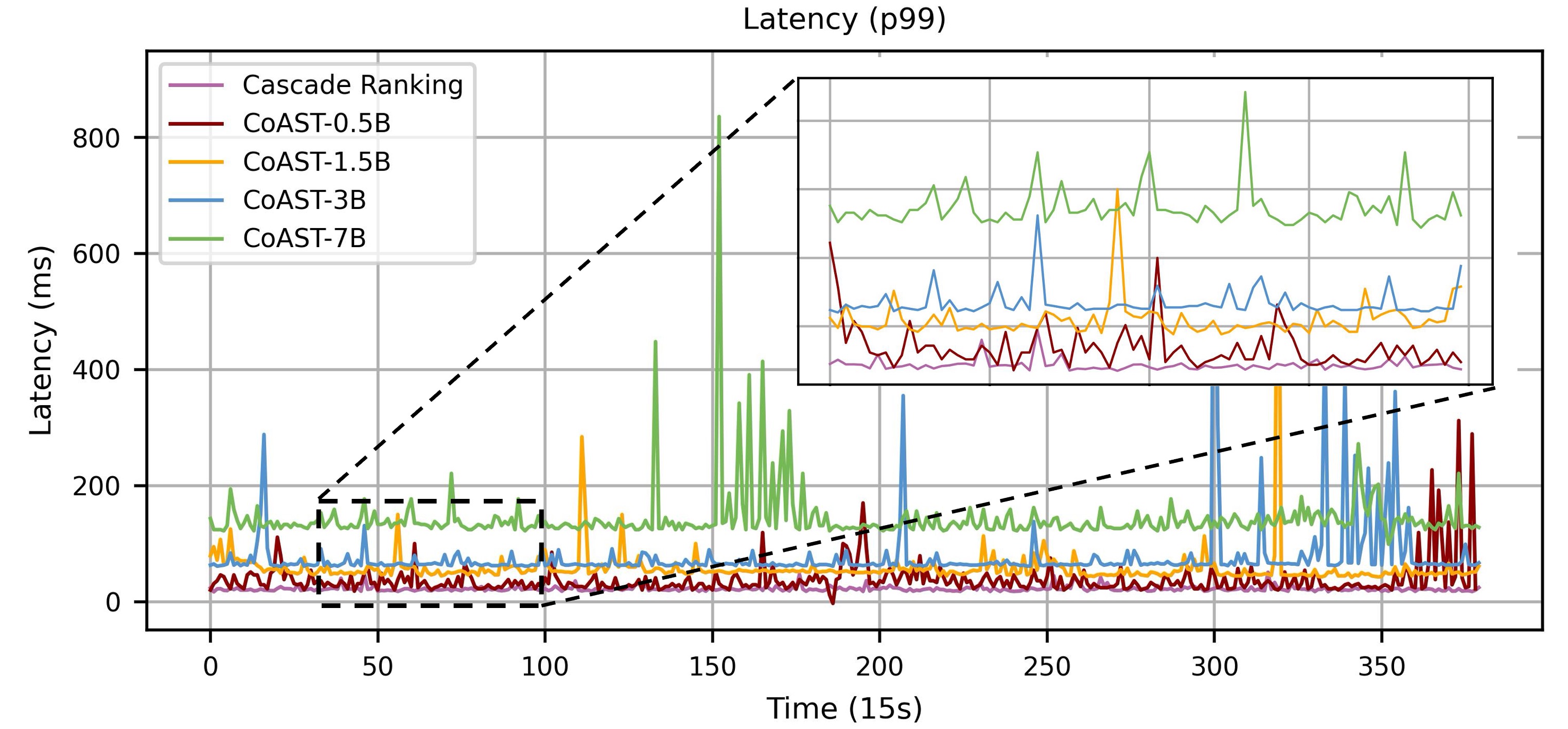}
    \caption{Comparison of Model Inference p99 Latency (ms) under a 50 QPS Load Test on 2$\times$ NVIDIA H20 GPUs.}
    \label{fig:load-test}
\end{figure}

To evaluate the online inference efficiency, we compare different variants of our model with the efficiency-SOTA cascade ranking system. Before deploying our model, we conduct a load test under the settings with a 50 QPS on 2$\times$ NVIDIA H20 GPUs.
Figure \ref{fig:load-test} presents the inference p99 latency results of different methods. As shown, the Cascade Ranking system, a widely adopted paradigm in industrial recommender systems, demonstrates exceptional efficiency with its latency consistently around 20ms. As a comparison, while CoAST-7B achieved the best offline performance among all our variants, its high computational cost posed a significant challenge for online serving. Even after applying the acceleration strategies described in Section \ref{sec:method}, its latency remained above 120ms, rendering large-scale deployment infeasible.

To strike a practical balance between efficacy and efficiency, we identified CoAST-0.5B as the optimal candidate. It delivers competitive performance while maintaining a manageable inference cost ($\sim$ 30ms). Consequently, we deployed CoAST-0.5B to production and conducted a online A/B test against the state-of-the-art (SOTA) Cascade Ranking baseline.

\subsubsection{Online A/B Test}
To assess the online effect of CoAST, 
we implemented a short-term and long-term A/B testing on "Guess Where You Go" card of AMAP's homepage. We examined CoAST-0.5B, trained based on a Qwen2.5-0.5B model, compared to the traditional cascade ranking baseline as shown in Table \ref{tab:online}. 

\begin{table}[h]
    \centering
    \setlength{\tabcolsep}{3pt}
    \caption{Online improvement over traditional Cascade Ranking baseline of short and long term A/B test. \% denotes relative improvement. short and long represents short-term and long-term A/B online test, respectively.}
    \label{tab:online}
    \begin{tabular}{lcc|ccc|c}
        \toprule
         &  P-CTR &  U-CTR &  SC-Rate & RE-Rate &   AC-Rate & NF-Rate \\
        \midrule
         short &  +5.06\% & +4.65\% &  +0.69\% &  +0.31\% &  +0.77\%  &   -5.11\% \\
         long &  +4.23\% &  +4.59\% &  +1.01\% &  +1.65\% &  +5.99\%  &  -5.85\% \\
        \bottomrule
    \end{tabular}
\end{table}

It is worth mentioning that this baseline already includes user profiles, trending topics and seasonal features for improving cognition in the cognitive re-ranking stage.
For short-term (one week) online test, CoAST demonstrate significant improvements in metrics like P-CTR (PV CTR) and U-CTR (UV CTR). This indicates that CoAST has increased the share of cognitive-aligned recommendations that effectively attracted user attention in terms of clicks. 
Additionally, it slightly enhances the user engagement metrics, such as SC-Rate (Scroll Rate), RE-Rate (User Retention Rate), and AC-Rate (User Active Rate), implying heightened user involvement. 
Notably, the effects on general utility metrics such as CTR and NF-Rate (Users' Negative Feedback Rate) are more modest, while the improvements on these engagement-related metrics often become apparent only over the long term.

To investigate the long-term effects of the proposed method in large-scale RSs, we maintain a small traffic for our baselines and compare it with CoAST for over one month. 
Compared to the baseline and the performance on the short-term test, our method has achieved a significant enhancement in the user engagement metrics, with a 5.85\% average decrease in NF-Rate (Users' Negative Feedback Rate). This demonstrates that CoAST could help to mitigate counter-intuitive recommendations and effectively improve user engagement while also boosting CTR, thereby enhancing the overall user experience.

\section{Conclusion}
\label{sec:conclusion}
In this paper, we proposed a LLM-based next POI prediction approach, named CoAST. 
In order to adapt LLMs to sequential recommendation tasks, we focused on a comprehensive pretraining task to adapt a base LLM to 
align language and collaborative semantics for recommendation. This task include two categories of corpus: sequential POI prediction and explicit index-language alignment.
Based on the learned item indices, our approach employed these alignment tasks to effectively adapt LLMs for sequential recommendation.
Furthermore, we give complete definition and metrics on cognitive-aligned next POI prediction task.
For constructing a cognitive-aligned POI recommendation system in industrial application, we design a multi-stage fine-tuning process, utilizing a curated dataset of cognitively-aware samples for both Supervised Fine-Tuning (SFT) and a subsequent Reinforcement
Learning (RL) phase. This procedure explicitly align the model with human common-sense.
Extensive offline experiments on three datasets demonstrated the effectiveness of our approach and Online A/B test achieves substantial improvements compared to traditional cascade ranking recommendation system.  

\begin{acks}
We would like to acknowledge the discussions of Jiawei Xue, Ning Wang, and Fangfang Chen throughout this work. Furthermore, we are grateful for  Tucheng Lin, Jian Song, Shulong Han, Jinhui Chen, Xiongfei Fan, Zudan Cao, Jing sun, Yifei Fan and Yukun Liu, for their engineering support.
\end{acks}

\bibliography{reference.bib}

\appendix

\begin{figure*}[h!]
    \centering
    \includegraphics[width=\linewidth]{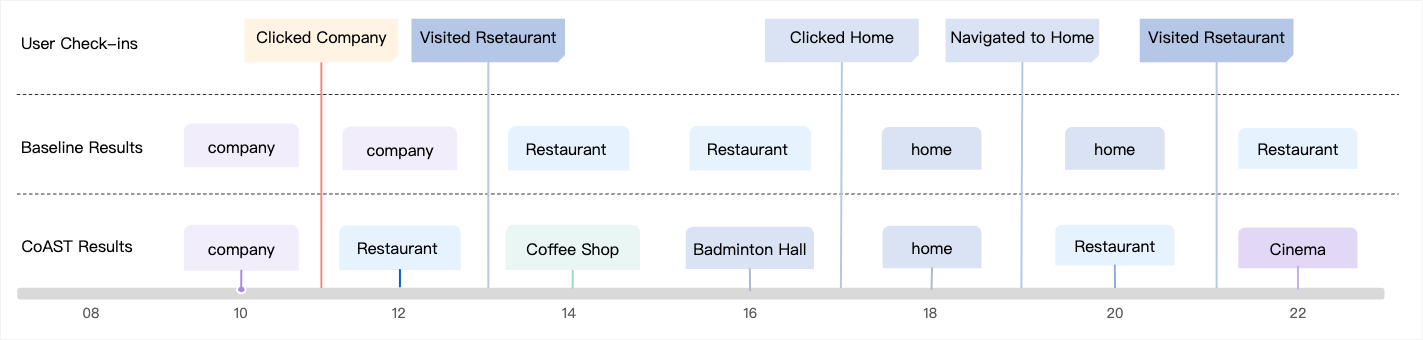}
    \caption{Case comparison of recommendation results of CoAST and the traditional basline RSs. The top row represents the user's real-time check-ins with different action types. The middle and bottom row represent the recommendation results of traditional basline RSs and CoAST, respectively.}
    \label{fig:case}
\end{figure*}

\section{Dataset}
\label{sec:dataset}
To validate our model's performance in a real-world application, we constructed a large-scale proprietary dataset by sampling desensitized users from AMAP (named Beijing). On this dataset, we conducted both extensive offline evaluations and an online deployment. Furthermore, to ensure a comprehensive and fair comparison against existing baselines, 
we also evaluate our approach on three real-world datasets: Foursquare-NYC \cite{6844862}, Foursquare-TKY \cite{6844862}, and Gowalla-CA \cite{10.1145/2020408.2020579}. 

The pre-processing follows the settings in \cite{wang2025generative}.
we pre-process each dataset by removing POIs with fewer than 10 interactions and users with fewer than 10 check-ins. 
The data is then sorted by time, with 80\% used for training, 10\% for validation, and 10\% for testing. Users and POIs that do not appear in the training set are removed from the test set to ensure consistency between training and evaluation. Subsequently, the check-in records are grouped by user and ordered chronologically. For each user, the last visited POI is held out as the ground truth during evaluation, while the preceding visits are used as the input
sequence. Note that in the training set, the user’s visit records are also treated as historical sequences and concatenated with the test set to the specified length before being fed into the model. 
Detailed dataset statistics are provided in Table \ref{label:stat}.

\begin{table}[h!]
\centering
\setlength{\tabcolsep}{5.pt} 
\renewcommand{\arraystretch}{1.2} 
\caption{Statistics of the preprocessed datasets. Avg.len denotes the average length of item sequences.}
\label{label:stat}
\begin{tabular}{lccccc}
\toprule[1pt]
\textbf{Dataset} & \textbf{\#Users} & \textbf{\#Items} & \textbf{\#Inter.} & \textbf{Avg.len} & \textbf{Sparsity} \\
\midrule
NYC & 1,083 & 5,135 & 104,074 & 136 & 98.1013\%   \\
TKY & 2,293 & 7,873 & 361,430 & 195 & 97.9798\%  \\
CA & 6,592 & 14,027 & 349,375 & 53 & 99.9996\%  \\
Beijing  & 10,000k & 1,520k & 38,0000k & 38 & 99.9999\%  \\
\bottomrule[1pt]
\end{tabular}
\end{table}

\section{Baseline}
\label{sec:baseline}
Following the settings of \cite{wang2025generative,zhang2024llm4poi}, we compare our proposed method with various next POI recommendation baselines, which can be specified as the following four categories:
(1) Traditional methods: PRME \cite{10.5555/2832415.2832536}, a ranking-based
metric embedding method; and PLSPL \cite{9117156}, a personalized sequential recommendation model. (2) Transformer-based methods:
STAN \cite{chen2023stan}, a spatio-temporal attention network; and GETNext \cite{10.1145/3477495.3531983},
a Transformer-based next POI recommendation model. (3) GCN-based method: STHGCN \cite{10.1145/3539618.3591770}, a spatio-temporal hierarchical graph
convolutional network. (4) Time-aware methods: TPG \cite{10.1145/3583780.3615083}, a timestamp-guided model; and ROTAN \cite{10.1145/3637528.3671809}, a time-aware POI recommendation framework. (5) LLM-based method: LLM4POI \cite{zhang2024llm4poi}, a method based on LLMs; and GNPR-SID \cite{wang2025generative}, a state-of-the-art based on Semantics codes and generative recommendation framework,
which is pretty closely related to our work.

\section{Evaluation Metric}
\label{sec:evaluation}
Following previous works, we employ the accuracy metric for the top-1 recommendation, denoted as \text{Acc}@1, at the specified time. 
\begin{equation*}
    \text{Acc}@1 = \frac{1}{n}\sum_{i=1}^{n}\mathbb{I}(y_i=\hat{y}_i)
\end{equation*}
where $n$ is the number of test samples, $y_i$ and $\hat{y}_i$ represent the
ground-truth label and the predict label of the $i$-th sample, respectively. $\mathbb{I}$ is the indicator function, which equals 1 when the condition is true and 0 otherwise.
Otherwise, we take the four cognitive categories (TCS, SCS, PAS and SAS) defined in \ref{sec:cognitive} as our metrics to evaluate the effect of the cognitive recommendation.

\section{Case Study}
\label{sec:casestudy}
Figure \ref{fig:case} illustrates the case comparison of recommendation results of CoAST and the traditional basline RSs.
In the traditional cascade ranking RSs, the model collapses to the user’s dominant daily rhythm: go to company at morning and back to home at afternoon. 
However, during other times of the day, conventional CTR-oriented models tend to be dominated by users' most recent interactions. This short-term focus often leads to recommendations that fail to capture the user's broader, long-term interests, resulting in a counter-intuitive user experience. 

In contrast, our proposed generative model achieves a better alignment of the user's temporal context with their holistic preference profile, thereby fostering a more well-rounded and sustainable recommendation ecosystem.

\end{document}